\documentclass[twocolumn]{article}
\usepackage[utf8]{inputenc}
\usepackage{cvpr}
\usepackage{times}
\usepackage{epsfig}
\usepackage{graphicx}
\usepackage{amsmath}
\usepackage{amssymb}
\usepackage{textcomp}
\usepackage{subfigure}

\usepackage[pagebackref=true,breaklinks=true,colorlinks,bookmarks=false]{hyperref}

\def\cvprPaperID{1234} % *** Enter the CVPR Paper ID here

\ifcvprfinal\fi

\cvprfinalcopy %un- comment this out

\title{Hierarchical Reinforcement Learning for Temporal Pattern Prediction}
\author{Faith Johnson}
\date{Sometime in the Future}

\begin{document}

\author{Faith Johnson\\
Rutgers University\\
{\tt\small faith.johnson@rutgers.edu}
\and
Kristin Dana\\
Rutgers University\\
{\tt\small kristin.dana@rutgers.edu}
}

\maketitle

\begin{abstract}
In this work, we explore the use of hierarchical reinforcement learning (HRL) for the task of temporal sequence prediction. Using a combination of deep learning and HRL, we develop a stock agent to predict temporal price sequences from historical stock price data and a vehicle agent to predict steering angles from first person, dash cam images. Our results in both domains indicate that a type of HRL, called feudal reinforcement learning, provides significant improvements to training speed and stability and prediction accuracy over standard RL. A key component to this success is the multi-resolution structure that introduces both temporal and spatial abstraction into the network hierarchy. 

\end{abstract}

\section{Introduction}

Reinforcement learning (RL) has made major strides over the past decade, from learning to play Atari games \cite{mnih2013playing} to mastering chess and Go \cite{silver2017mastering}. However, RL algorithms tend to work in a specific, controlled environment and are often difficult to train. In response to this brittleness, hierarchical reinforcement learning (HRL) is growing in popularity.  

We combine deep learning and HRL for temporal sequence prediction in two application domains where publicly available data is abundant. First, we develop stock agents to execute trades in a market environment. The training data consists of historical stock prices from 1995 to 2018. Second, we develop a vehicle agent to predict steering angles given visual input. We use the Udacity dataset\cite{udacity} as training data, which consists of five videos with a total duration of 1694 seconds.  

In HRL, a manager network operates at a lower temporal resolution and produces goals that it passes to the worker network. The worker network uses this goal to produce a policy over micro-actions at a higher temporal resolution than the manager\cite{vezhnevets2017feudal}. Within the stock market, there are two natural hierarchies. The first hierarchy involves temporal scales. A trader can consider how a stock's price fluctuates over the course of an hour, but also over the course of a week, month, or year. The second hierarchy is the separation of the different market sectors, each containing a multitude of stocks. In order to trade in the market effectively, stock brokers must consider both the relationships between sectors and the relationships between the stocks in each sector. 

In the same vein, the task of autonomous navigation is complicated because, at all times, human drivers have two levels of things they are paying attention to. The first level is on a fine grain: don't immediately crash the vehicle by hitting obstacles. The second level is on a coarser grain: plan actions a few steps ahead to keep the vehicle going in the correct direction as efficiently as possible. 
%These breakdowns into a low level and high level task lend themselves to a hierarchical reinforcement learning approach. 
In both domains, financial and vehicluar, we implement agents with both RL and HRL. We show that HRL provides improved training stability and prediction performance.

\section{Methods}

\subsection{LSTM Stock Predictions}
First, we set up a baseline for future result comparison using a simple LSTM network. Using stock market data gathered from Kaggle\cite{kaggleStocks}, we predict the closed price for a single day given the open price of that day. We build a simple LSTM model in Keras\cite{chollet2015keras} with ten neurons and a ReLU activation function followed by a fully connected layer. This network is trained for ten epochs using the mean squared error loss function and an adam optimizer. 

Next, we predict a sequence of open prices for a particular stock given a sequence of previous open prices. For this, we use a slightly larger LSTM network with three layers of LSTMs, each with ten neurons and ReLU activation functions followed by a fully connected layer. The loss function and optimizer for this experiment are also mean squared error and Adam respectively. However, this network is trained for twenty epochs. We conduct this experiment with several sequence pairings: 1 previous price to predict the next price, 3 previous prices to predict the next 3 prices, and 5 previous prices to predict the next 5 prices.

Finally, we implement a reinforcement learning stock agent to predict open stock prices by learning a multiplier to transform the previous open price to the next open price. It takes a sequence of historical open prices for a single stock as input and passes them through several layers of LSTMs with ReLU activation functions and a fully connected layer. The new price is computed by multiplying the current open price by the output to produce the predicted open price for the next day.

\subsection{Stock Environment}
After the multiplier agent, we rethink our approach to stock price prediction by moving away from predicting the price itself, and thus away from computing the answer to a regression problem. The next set of experiments involves executing trades in a stock market environment with the goal of doubling the value in a given portfolio. The rationale behind this change is that reinforcement learning excels at learning a policy over a given set of actions. In the regression problem of learning a multiplier, the action space is essentially infinite. In addition, the state space is infinite because it consists of all possible stock prices. Learning a policy over an infinite set of possibilities is almost impossible. 

To create our stock environment for these new reinforcement learning agents, we use Quadnl\cite{QuandlWIKI} to collect market information for a small subset of stocks from six sectors. The six stock sectors are technology, energy, finance, healthcare, utilities, and transportation. The goal is to have a diverse selection of stocks and sectors from the market in order for the agent to be able to glean the relationships between the sectors along with the relationship between stocks within each sector.

We define the action space to consist of three actions: buy, sell, and hold. All agents buy or sell only one stock per action, unless otherwise stated. If an agent does not have enough money to execute the buy action, it is forced to hold. The same is true if it does not have enough shares to execute the sell action. The environment keeps track of an agent's current balance of cash and portfolio value, as well as giving the agents a reward for their actions, which is generally the change in the agent's total portfolio value.

\subsection{Hard Coded Stock Agent}
The baseline for performance in the stock environment comes from a hard coded stock agent whose aim is to double the value of a given portfolio. First, it defines two thresholds, a selling threshold and a buying threshold. When two consecutive open prices differ by less than the selling threshold, the agent decides to sell shares in the stock. When the two prices differ by more than the buying threshold, the agent decides to buy a share in the stock. The idea is that the buying threshold is positive and the selling threshold is negative so that the agent will sell when the price goes down and buy when the price goes up. When the price difference lies between the two thresholds, the agent takes the hold action.

\subsection{Reinforcement Learning Stock Agents}

\subsubsection{Q Learning Agent}
For the Q learning agent, the state space is limited to the combinations of whether or not the price of a stock has gone up or down and whether or not the agent currently possesses shares in said stock. In q learning, the agent keeps track of the policy over actions and states using a q table. The q table is indexed by the actions and the states, as in Table \ref{table:QTableStock}, and is initially filled with zeros. As an agent takes actions in the environment, it receives a reward that it uses to update the table to reflect the utility of each action given a certain state. An agent decides which action to take by referencing the portion of this q table corresponding to its current state and choosing the action associated with the highest q value.

\begin{table}[h]
\centering
\begin{tabular}{ |c||c|c|c| }
\hline
 & Buy & Hold & Sell \\
\hline
\hline
Price Increases \& Have Shares &  &   &   \\ 
\hline
Price Increases \& Have No Shares &  &   &   \\ 
\hline
Price Decreases \& Have Shares &  &   &   \\ 
\hline
Price Decreases \& Have No Shares &  &   &   \\ 
\hline 
\end{tabular}
\smallskip
\caption{Example q table for the q learning stock agent. It is indexed by the three actions (buy, hold, sell) and the combination of price fluctuation and share possession. }
\label{table:QTableStock}
\end{table}

\subsubsection{Deep Q Network (DQN) Agent}

This stock agent builds upon the same ideas as the previous q learning agent, but it chooses it's actions differently. Instead of using the reward from the environment to update the q table, it uses a deep q network, or DQN, to approximate the q value of a certain action given a state. This network is made up of three LSTMs with ReLU activation functions in sequence, followed by a fully connected layer. The first LSTM has 32 hidden layers, and the last two have 64 layers. 

In the case of this network, the state is the previous three open prices for the stocks in each of the sectors. However, the action space remains the same. With the q learning agent, having an infinite state space would not be ideal for optimal policy convergence, but the DQN is still able to converge on a solution.

The reward from the environment plays a role in the loss back-propagation of the network. For each action taken by the agent, a tuple containing the initial state, $s_0$, the final state, $s$, the action taken, $a$, and the corresponding reward, $r$, is saved in a replay buffer. During training, a random tuple is sampled from this buffer, and the loss, which in this case is the reward, is back-propagated through the network as in \cite{mnih2013playing}.

\subsection{Feudal Reinforcement Learning}
In feudal reinforcement learning, the manager network operates at a lower temporal resolution than the worker network. It receives state input from the environment and communicates with the worker network through a goal vector. This goal vector encapsulates a temporally extended action that the manager thinks will receive the highest reward from the environment. The worker executes atomic actions in the environment based on this goal vector and its own state information. This process of manager/worker communication through temporal abstraction helps to break down a problem into more easily digestible pieces.

To explain the concept of temporal abstraction further, take the case of an agent attempting to leave a room through a door. When a person thinks of completing this action, they don't do it at the low level of straight, straight, left, straight, right, etc. In other words, they do not consciously think of each atomic action required to exit the room. Instead, they think in terms of temporal abstraction. Find the door. Approach it. Pass through it. Each of those actions encapsulates multiple atomic actions that need to be executed in a specific order for the agent to complete the task.

For a feudal network to solve the room example, the manager would create goal vectors for the ``find the door", ``approach it", and ``pass through it" operations. Then, the worker would only have to focus on executing atomic actions to complete one of these smaller tasks at a time, which is much simpler than the original task of  exiting the room as a whole. This makes it easier to generate an ideal policy. Additionally, the idea of temporal abstraction can be applied to space. Incorporating different spatial resolutions into a feudal network can  break down problems into spatial abstractions which make them easier to solve in the same way.

\subsection{Maze Environment}
To test the performance of feudal reinforcement learning, we use a maze environment as proposed in Dayan et al.\cite{dayan1993feudal}. In this environment, there are multiple levels of the same maze, each at a lower spatial resolution than the previous level. The agent at the highest spatial resolution is the worker, who receives goal vectors from the agent at the next lowest resolution, who is its manager. This manager becomes the worker for the agent at the next lowest resolution, and so on, until you reach the level with the lowest spatial resolution, where the ultimate manager resides. For example, the worker on the level with the highest spatial resolution will operate in a 16x16 grid and have a manager who operates in an 8x8 grid. This manager will be the worker for the agent in the 4x4 grid, etc., until you get to the final manager in the 1x1 grid. 

\begin{figure}[]
    \centering
    \subfigure[Manager View]{\includegraphics[width=.45\linewidth]{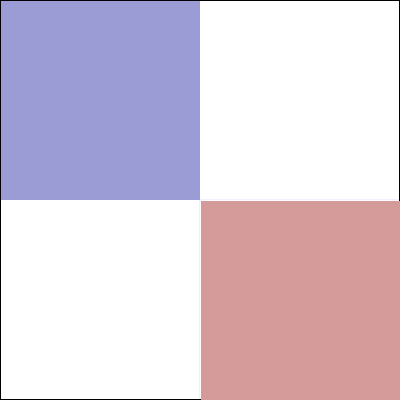}}
    \subfigure[Worker View]{\includegraphics[width=.45\linewidth]{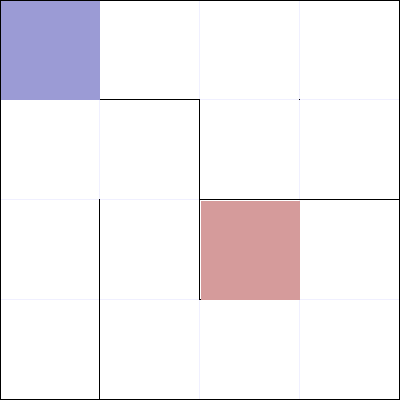}}
    \caption{(a) Manager's 2x2 view of the maze. (b) Worker's corresponding 4x4 view of the same maze.}
    \label{fig:mazeViews}
\end{figure}

We create this maze by editing the gym-maze\cite{mazeEnv} github repository code. In our maze, there are only two levels. The worker operates in a 4x4 version of the maze, and the manager operates in a 2x2 version, as in Figure \ref{fig:mazeViews}. Each square in the manager's 2x2 rendition of the maze corresponds to a 2x2 section of the worker's maze. We omit the 1x1 manager from the Dayan et al. experiment because it is computationally irrelevant for this task. The state space of each agent is comprised of each square in the grid of their respective maze resolutions. The objective of the agents is to reach some goal square in the maze. This goal is mapped to the same equivalent location in all maze levels and can be specified at run-time. The action space is comprised of moving north/south/east/west or declaring that the goal is the current space. There is a base reward of \[ \frac{-0.1}{X_{dim} * Y_{dim}} \] applied to every movement for all agents, where$X_{dim}$ and $Y_{dim}$ are the x and y dimensions of the maze. However, if an agent finds the goal, it receives a reward of 1.

\subsection{Maze Agents}
We test several agents in our maze environment, starting with a reinforcement learning model as a baseline, before moving to a feudal reinforcement learning implementation. 

\subsubsection{Q Learning Agent}% and DQN Agents}
The first agent we built for the maze environment uses q learning to navigate a single, 4x4 level of the maze in search of a goal square. When it reaches this goal, the experiment ends. The q table of the agent is the same size as the maze with each square in the maze corresponding to one entry in the table. The values in this table are updated based on the reward received from the environment.

%We also implement a DQN maze agent that models the q table using a deep neural network made up of three LSTMs with ReLU activation functions in sequence. The first LSTM has 32 hidden layers, and the last two have 64. This is followed by a fully connected layer and is trained using a replay buffer.

\subsubsection{Feudal Q Learning Agent}
The feudal network solves the maze using q learning as well. The manager network receives it's location in the maze as it's current state. It uses this to choose and execute the best action from it's q table. The manager is able to move in any of the four directions, and this direction is the basis for the goal vector that tells the worker what quadrant to move to. If the manager declares that the goal is in its current space, it is telling the worker that it should look for its own goal in a specific quadrant of the maze. When the worker receives a goal vector from the manager, it moves to the specified quadrant and waits for more instructions. If it is indicated that the worker should look for the goal in a quadrant, it continues to move until the goal is found.

Both the manager and the worker receive a base negative reward for every action that doesn't result in finding the goal. In this way, the spaces furthest away from the goal will have a more negative q value than those closest to the goal. In addition, the manager receives a negative reward if the worker finds the goal space without following the instructions from the goal vector. While the individual reward values resulting from exploring the maze may be the same, the manager and worker do not receive the same reward signals. The worker takes many more steps in the environment due to the difference in spatial resolution, so it will receive a reward more often than the manager. In this way, the spatial abstraction of the maze results in a temporal abstraction of the reward signals.

\subsection{Feudal Reinforcement Learning Stock Agents}
Once we discovered the performance improvements of feudal reinforcement learning, we decide  to return to the stock market portfolio experiments with feudal reinforcement learning agents.

\subsubsection{Feudal Q Learning Agent}
Our feudal q learning stock agent operates in the same environment as the previous reinforcement learning agents and has the same goal of doubling the value of some portfolio. Its input is a sequence of open prices for each stock in the six predetermined sectors. The q table structure and state space are also the same. The main difference is the division of labor between the manager and worker networks. 

The manager receives the price input from the environment and determines whether or not each of the six sectors should be traded or not. This decision is passed to the worker in the goal vector. The worker then decides whether to buy or sell the stocks in the sectors specified by the manager. For each goal vector from the manager, the worker acts a fixed number of times to introduce temporal abstraction to the problem in addition to the spatial abstraction already present. The reward of the manager is the overall portfolio value change, while the worker receives a reward for the portfolio value change of each sector after each action it executes. 

% \subsubsection{Feudal DQN Agent}
% We also implement a feudal deep q network agent to perform the stock experiments that works much the same way as the feudal q learning agent. Instead of updating a q table, both the manager and worker have deep neural networks to estimate the q value of a particular state and action. These networks are updated using replay buffers. 

\subsubsection{Feudal Networks with Multiple Workers}
The feudal reinforcement learning problem can be extended to a vertical hierarchy with multiple managers and workers in sequence, as we've already explored, but this concept can also be extended horizontally to one manager with multiple workers. To this end, we have implemented two different experiments, both using q learning. The first involved a manager network with a set of three different worker networks, where each worker makes a different number of transactions in the environment. The manager's action is choosing which of the workers will act in the environment at a given time.

The second involved a manager network with a set of three workers, where each of these workers have a different hard coded behavior. The first buys when a stock's price increases and sells when it decreases, the second sells when a stock's price increases and buys when it decreases, and the third executes a random action. The manager's action set consists of choosing which worker will interact with the environment.

\subsection{Driving Environment}
We also test feudal reinforcement learning in the domain of autonomous vehicles. For that, we use the Udacity driving dataset\cite{udacity}. They provide steering angles, first-person dash cam images, braking, and throttle pressure data. We augment this dataset to increase its size and influence model training by performing several transformations on the image and angle data. First, we implement a horizontal flip to effectively double the size of the dataset. For this change, we negate the angles associated with the flipped images. As an additional option, we use the horizontal and vertical optical flow images. The horizontal optical flow image, $i_x$, is obtained by convolving the image with the row vector $[ 1, 0, -1]$, while the vertical optical flow image is obtained by convolving the image with the column vector \[\begin{bmatrix}
1 \\
0 \\
-1 \\
\end{bmatrix}\] Finally, all images are scaled and normalized so that their pixel values lie in the range $[-1,1]$.

\subsection{Steering Angle Experiments}

\subsubsection{Steering Angle Prediction}
We started simple, so our first task was to predict steering angles based on visual input. After some initial difficulty with our model, we found a network\cite{komanda} from the Udacity challenge that accurately predicts steering angles. It has a convolutional layer with a ReLU activation function followed by a dropout layer. The output of this is saved to use for a skip connection later on in the network. This is repeated four times before the output is fed through some fully connected layers, also with ReLU activation functions. At this point, the output and the intermediary representations are added together, passed through an ELU layer, and normalized. Then, the previous steering angle and the output of the ELU layer are passed through an LSTM. Finally, the output of the LSTM is passed through a fully connected layer to produce the steering angle. Note that this network takes in a sequence of images as well as the previous angle in order to make its predictions.

\subsubsection{Subroutine ID Prediction}
Being able to predict steering angles is useful, but for feudal reinforcement learning we also need to classify the steering angles into their temporally abstracted categories (such as go right, go left, go straight). This can be done by hand, but it would be a lengthy process. Instead, we take inspiration from Kumar et al.\cite{kumar2019learning} to learn these subroutines, otherwise called options or macro-actions, using a neural network. 

To do this, we jointly train two networks. The first takes in a sequence of angles and predicts the subroutine ID. The second takes in the subroutine ID, a sequence of images, and the previously predicted angle and predicts the next steering angle in the sequence. A problem we encountered with the steering angle prediction from the previous section is that it appears as if the network is simply predicting that the previous angle will be the next steering angle. To circumvent this, we give the second network the previously predicted angle instead of the ground truth angle. Additionally during training, the sequence of angles fed into the first network contains the angle it is trying to predict. However, during testing, we only use a sequence of angles preceding the angle we aim to predict in order to avoid this conflict.

\subsubsection{t-SNE Prediction}
\begin{figure}[]
    \centering
    \includegraphics[width=.9\linewidth]{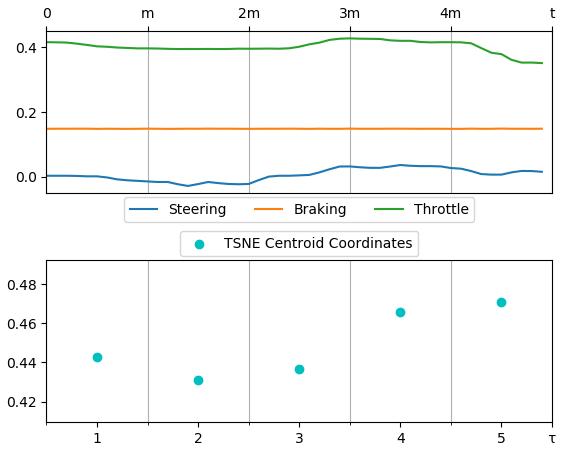}
    \caption{Steering, braking, and throttle data are concatenated every m time steps to make a vector of length 3m. Each 3m vector corresponds to one set of coordinates in the 2D t-SNE space. The t-SNE coordinates act like a manager for the steering angle prediction and operate at a lower temporal scale. In our experiments, m=10. $t$ and $\tau$ correspond to the final time step for the driving data and t-SNE coordinates respectively.}
    \label{fig:tsneBreakdown}

    \centering
    \includegraphics[width=.9\linewidth]{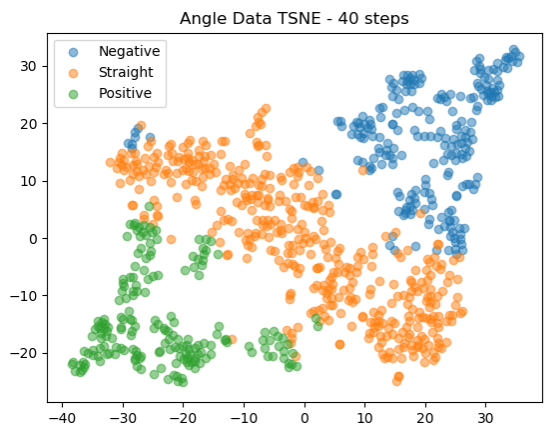}
    \caption{Total plot of the t-SNE coordinates for the Udacity data. The colors correspond to the average sign of the angles in each length 3m vector used to generate the points.}
    \label{fig:tsnePlot}

    \centering
    \includegraphics[width=.9\linewidth]{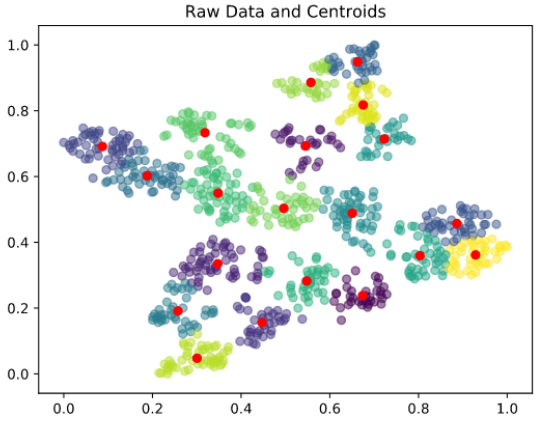}
    \caption{K-Means clustering (k=20) of the TSNE coordinates of the Udacity data with the centroids pictured in red. Not only to distinct clusters form in the data, but each cluster corresponds to a unique action of the vehicle.}
    \label{fig:kmeansCentroids}
\end{figure}

\begin{figure*}[]
    \centering
    \includegraphics[width=.99\linewidth]{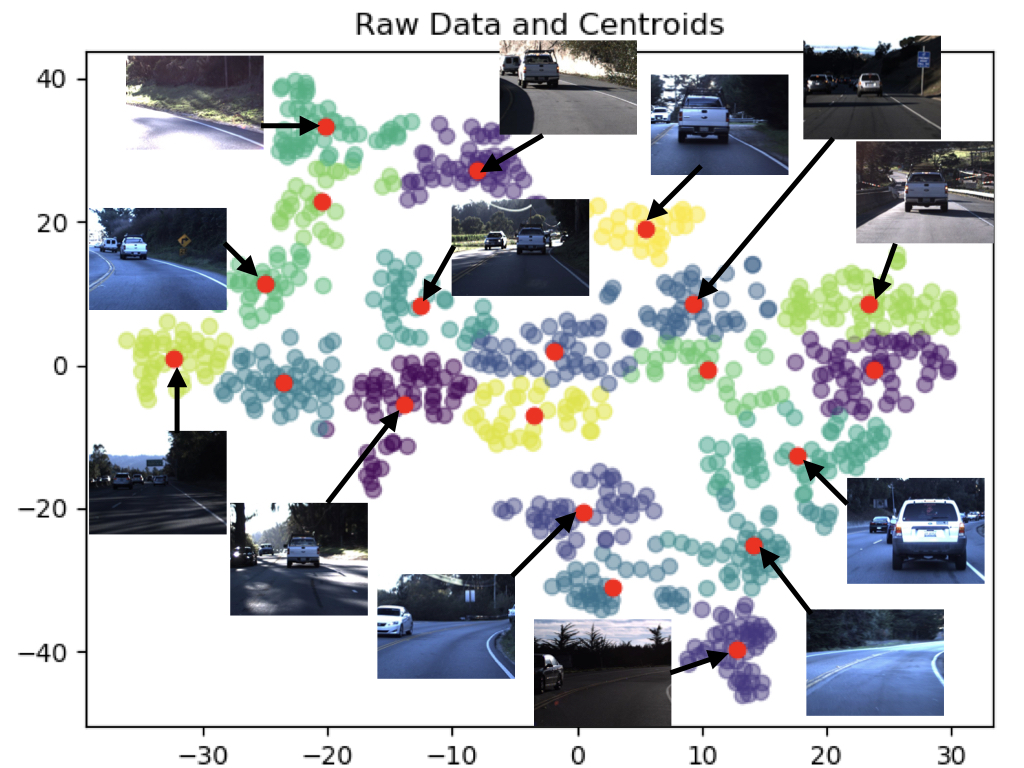}
    \caption{Example training images are shown with their corresponding t-SNE centroids. Notice that the bottom right of the figure contains sharp right turns. As you move upwards, the right turn gets less sharp until the vehicle begins to go straight. By the top left of the figure, the vehicle is making sharp left turns. }
    \label{fig:tsnePhotobomb}
\end{figure*}

Ideally, we want an angle prediction network that does not take in the previous steering angle at all. To accomplish this, we explored using t-SNE\cite{maaten2008visualizing} as an embedding space for our driving data and as the subroutine IDs themselves. To do this, we arranged the steering angle, braking, and throttle pressure data into vectors of length ten. Then, the vectors from each category that correspond to the same time steps are concatenated together to make vectors of length thirty. The collection of these vectors is passed through the unsupervised t-SNE algorithm to create a coordinate space for the driving data. 

Each vector of length thirty is given one x and y coordinate pair as illustrated in Figure \ref{fig:tsneBreakdown}. The greater collection of all of the generated points is in Figure \ref{fig:tsnePlot}. The coloring of the points in this figure is hard coded. The points corresponding to vectors with primarily negative steering angles are in blue. The points corresponding to vectors with positive steering angles are in green. The orange points correspond to vectors with steering angles that are relatively close to zero. 

Once we have the t-SNE embedding of the data, we use K-Means clustering on the coordinates and take the centroids of the clusters as our new subroutine IDs, as shown in Figure \ref{fig:kmeansCentroids}. We vary k from ten to twenty to determine if different numbers of clusters improve prediction performance. Then, we train a network to take in the centroids as the subroutine ID, as well as a sequence of images, in order to predict the next steering angle. 

In order to ensure that no data pertaining to the predicted steering angle is used as input to this network, we use the t-SNE centroid corresponding to the data of the previous 3m steering, braking, and throttle data as input to the network. To illustrate, refer back to Figure \ref{fig:tsneBreakdown}. If we are predicting an angle from the range $t\in[2m,3m]$, then the t-SNE centroid used for the subroutine ID input to the angle prediction network will be the centroid at $\tau = 2$, which was made with the steering, braking, and throttle data from $t\in[m,2m]$. In this way, the angle we are attempting to predict will not be used to compute the t-SNE centroid used as the subroutine ID. This shift also incorporates an extra level of temporal abstraction into our network.

Additionally, we create a tool that displays the visual data corresponding to the different t-SNE coordinates, allowing the user to visually inspect that neighboring points in the embedding space correspond to similar driving behaviors. Figure \ref{fig:tsnePhotobomb} attempts to replicate this by showing example training images that correspond to some of the t-SNE centroids. Notice that the bottom right of the figure contains sharp right turns. Moving diagonally upwards, the right turns get less sharp until the vehicle begins to go straight. Then, this straight motion gradually begins to become a left turn until, by the top left of the figure, the vehicle is making sharp left turns. 

\section{Results}
% \begin{itemize}
    % *** restart the date vs don't restart the date ***
% \end{itemize}

\subsection{LSTM Experiments}
\begin{figure}[]
    \centering
    \includegraphics[width=.9\linewidth]{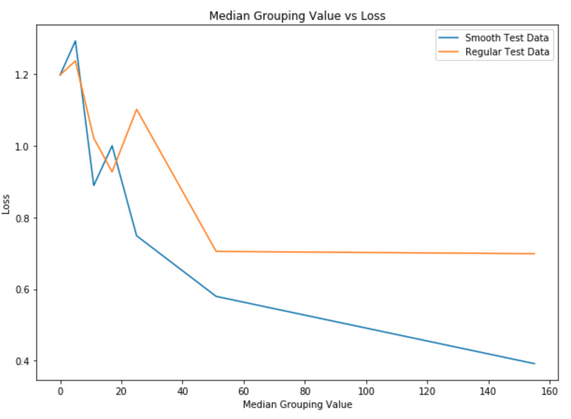}
    \caption{LSTM loss comparison for predictions on regular data and smoothed data. The more smoothed the data, the more quickly the loss decays.}
    \label{fig:lossComp}
\end{figure}

\begin{figure*}[]
    \centering
    \subfigure[Window of 2]{\includegraphics[width=.49\linewidth]{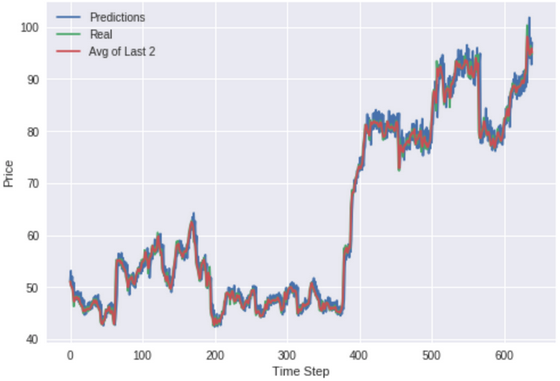}}
    \subfigure[Window of 4]{\includegraphics[width=.49\linewidth]{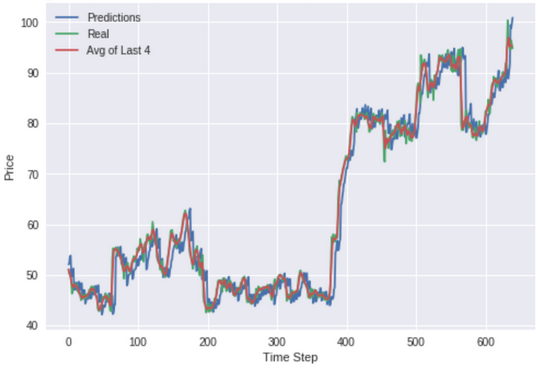}}
    
    \subfigure[Window of 10]{\includegraphics[width=.49\linewidth]{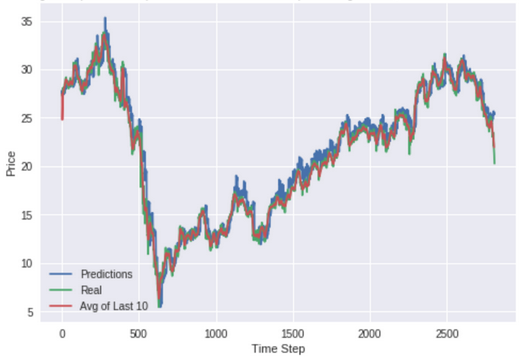}}
    \subfigure[Window of 12]{\includegraphics[width=.49\linewidth]{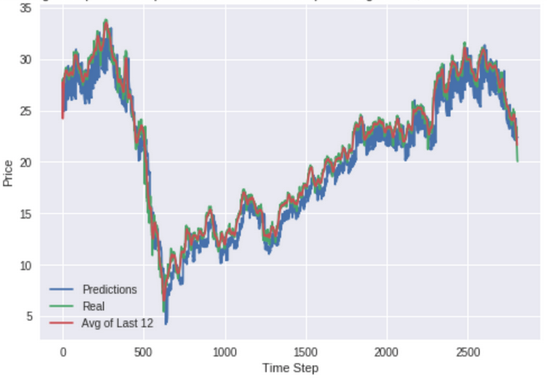}}
    \caption{Prediction results for an LSTM when the size of the input and output prediction windows are from two, four, ten, and twelve respectively. The LSTM predictions match more closely with smaller windows than larger windows and provide less noisy results. However, the larger windows allow for longer term predictions.}
    \label{fig:windowTrials}
\end{figure*}

Our first experiment used LSTMs to predict open prices of stocks. We varied the input and output window sizes from one to fifteen and compared the results. A subset of the prediction graphs are available in Figure \ref{fig:windowTrials} for predicting two, four, ten, and twelve prices out. Also included in each graph is a line representing the average of the last two, four, ten, or twelve prices as a comparison to the prediction. The prediction with a window of two works extremely well, as evidenced by the fact that the three lines (the real, prediction, and average price) are almost directly on top of each other. However, as the window length increases, there is a clear divergence of the prediction from the real price.  There is a trade off between accuracy and the length of the prediction, which is expected because data farther out in time will have less of a dependence on the input to the LSTM. Also, the predictions become much noisier, which is to be expected for the same reason.

We also tested LSTM prediction on smoothed data. After training the LSTM on this smoothed data, we tested the model on new smoothed and non-smoothed data and compared the loss values, computed through mean squared error, in Figure \ref{fig:lossComp}. The blue line is the loss associated with the smoothed data and the orange line corresponds to the regular data. The loss value is on the y-axis, and the x-axis shows how much smoothing was applied to the training data. We smoothed using a moving average filter, so the x-axis points represent how many data points were used in the average. The graph shows that feeding smoothed data into an LSTM increases the accuracy of its predictions and leads to a quicker decay of the loss.

\begin{figure}[]
    \centering
    \subfigure[Real (blue) versus predicted (green) stock price predicted using the multiplier stock agent]{\includegraphics[width=.9\linewidth]{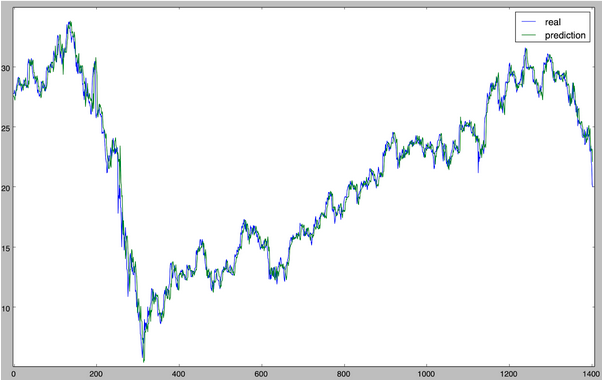}}
    
    \subfigure[Difference between the real stock price and the predicted price for the multiplier stock agent]{\includegraphics[width=.9\linewidth]{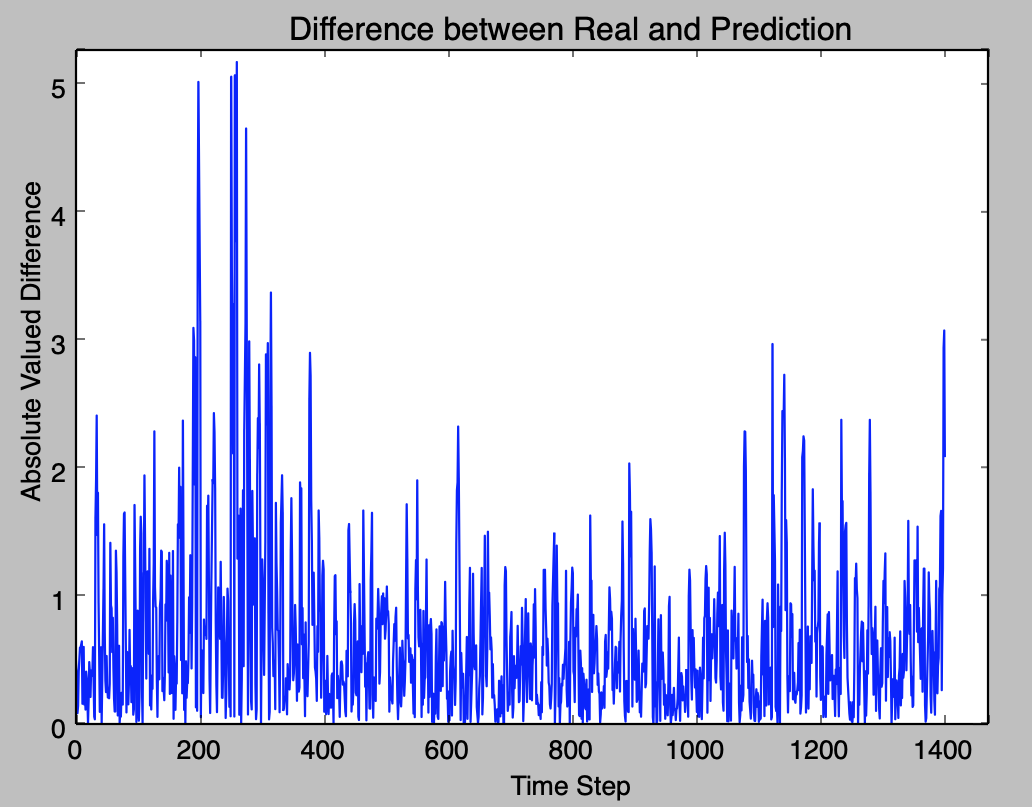}}
    
    \caption{Results for the multiplier stock agent. (a) shows that the predictions match very closely, and (b) shows that the areas where the predicted and real price differ the most occur during drastic price changes.}
    \label{fig:multAgentDiffPreds}
\end{figure}

The results for our first stock agent that learns a multiplier to predict the next open price based on the previous stock price can be found in Figure \ref{fig:multAgentDiffPreds} (a). The blue line represents the real stock open price, and the green line is the prediction. The y-axis is the price in dollars, and the x-axis is the time step. To get a better idea of exactly how accurate the predictions are, Figure \ref{fig:multAgentDiffPreds} (b) also shows the difference between the real open prices and the predicted prices for each time step. Most of the predictions differ by less than a dollar, which is an order of magnitude less than the prices themselves. The largest differences between the real and predicted prices occur during drastic changes to the stock price. 

\begin{figure}[]
    \centering
    \subfigure[Multiplication RL Stock Agent]{\includegraphics[width=.9\linewidth]{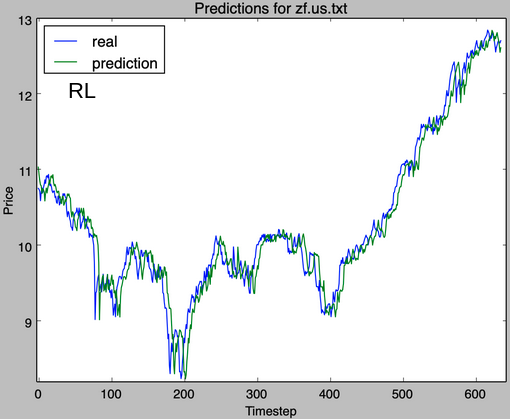}}
    \subfigure[LSTM]{\includegraphics[width=.9\linewidth]{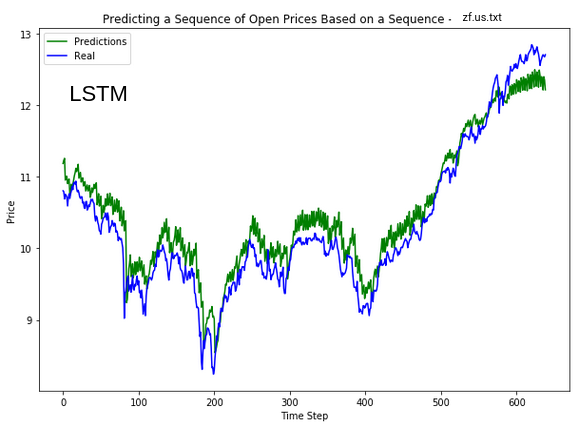}}
    \caption{Comparison of the multiplication stock agent predictions and the LSTM predictions. The multiplication agent matches the real prices closer than the LSTM and provides less noisy predictions overall.}
    \label{fig:RLvsLSTM}
\end{figure}

We then compare these predictions with those from the LSTM. In general, the reinforcement agent predictions, as pictured in Figure \ref{fig:RLvsLSTM} (a), are much more accurate than the LSTM prediction, as pictured in Figure \ref{fig:RLvsLSTM} (b). It seems that LSTMs do a decent job at predicting smaller, local changes, but their performance falls short when a large change in price occurs. The reinforcement learning agent is more robust to being able to handle these changes. Additionally, the LSTM predictions are much more noisy.

\begin{figure}[h]
    \centering
    \subfigure[Multiplier Agent Predictions]{\includegraphics[width=\linewidth]{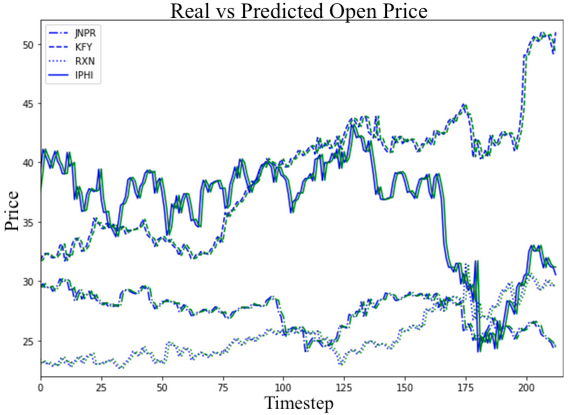}}
    
    \subfigure[LSTM Predictions]{\includegraphics[width=\linewidth]{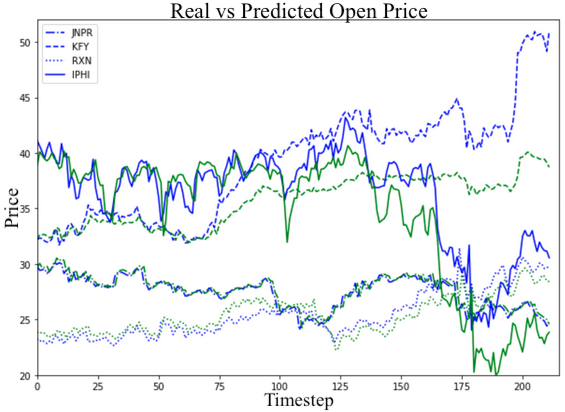}}
    \caption{Comparison between the multiplier stock agent and an LSTM for predicting prices for multiple stocks at once. The stock symbols are provided in the legend (top left). The LSTM struggles to capture the behavior of multiple stocks at once. }
    \label{fig:compForMultStocks}
\end{figure}

We also compare the LSTM and multiplication stock agent's abilities to predict open prices for multiple stocks at once in Figure \ref{fig:compForMultStocks}. Once again, we see that the predictions of the reinforcement learning stock agent are much better than the LSTM. For this experiment, we used stocks from the same sector in order to increase the likelihood that there would be correlations in the stocks' behavior. Reinforcement learning is better able to find and exploit these relationships to help make predictions than the LSTM. With this in mind, we shift our focus towards reinforcement learning and other techniques that can be derived from it.

\subsection{Maze Experiments}

\begin{figure}[h]
    \centering
    \includegraphics[width=\linewidth]{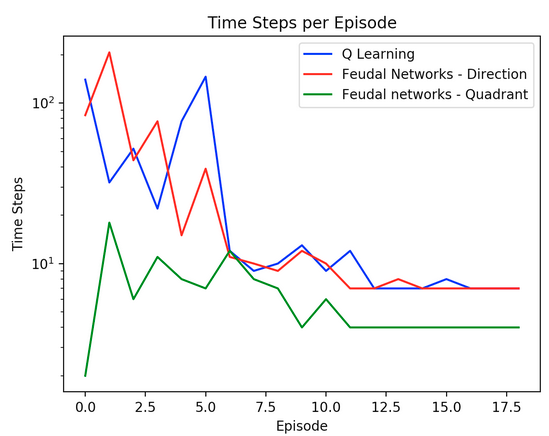}
    \caption{Comparison of time steps required per episode for three agents in the maze environment. Feudal reinforcement learning (green) takes less time to solve the maze overall and explores the maze more efficiently.}
    \label{fig:feudalShorter}
\end{figure}

\begin{figure}[h]
    \centering
    \includegraphics[width=\linewidth]{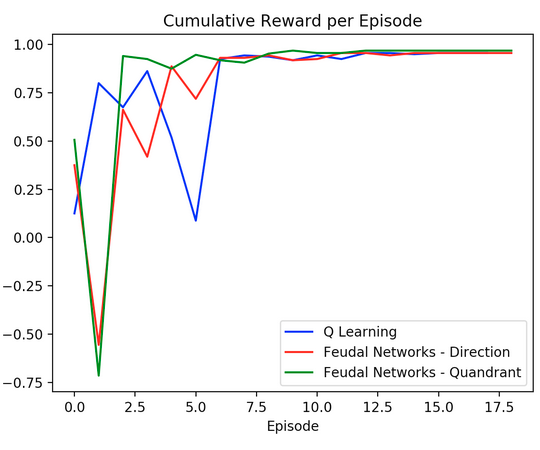}
    \caption{Comparison of the reward per episode for three agents in the maze environment. The reward for feudal reinforcement learning (green) converges faster than the other two methods.}
    \label{fig:feudalHigher}
\end{figure}

Moving to the maze experiments, we compare the relative performance of reinforcement learning and feudal reinforcement learning. We had three different agents navigate a maze until they reached some goal space. For this experiment, we used a fixed maze with a fixed goal space, but it is possible to randomize both the maze structure and the goal location at run time. The standard q learning agent's performance is documented in Figure \ref{fig:feudalShorter} and Figure \ref{fig:feudalHigher} in blue. 

The next two agents are different variations of feudal q learning networks. In the first, the goal vector the worker receives from the manager tells it which direction to take in the maze. The results for this agent are in red in Figures \ref{fig:feudalShorter} and \ref{fig:feudalHigher}. In the second feudal q learning agent, the goal vector received by the worker tells it which quadrant to go to in the maze. This is different than the previous agent because the worker is not explicitly told which direction to take to reach this goal quadrant. This agent's results are pictured in Figures \ref{fig:feudalShorter} and \ref{fig:feudalHigher} in green.

In Figure \ref{fig:feudalShorter}, we compare the amount of time steps per episode for each of the three agents. The q learning agent takes the most amount of time overall, closely followed by the feudal network with a direction as the goal vector. Additionally, these two methods solve the maze in approximately the same amount of time once they have found the optimal path through the maze. However, the feudal network with a quadrant as the goal vector is both significantly faster during training and finds a better solution to the maze, as evidenced by the fact that its solution takes less time to navigate the maze than the other two agents. 

\begin{figure*}[]
    \centering
    \subfigure{\includegraphics[width=.246\linewidth]{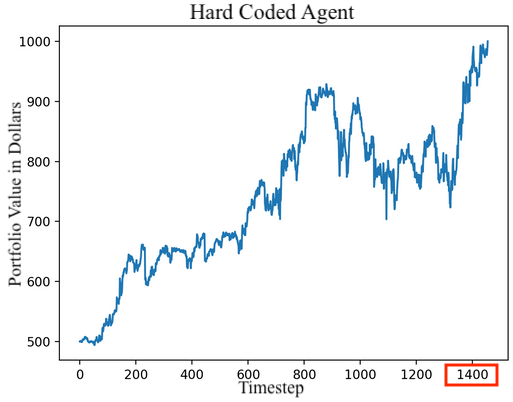}}
    \subfigure{\includegraphics[width=.246\linewidth]{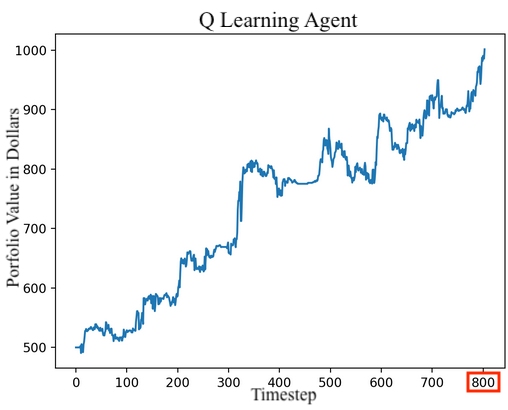}}
    \subfigure{\includegraphics[width=.246\linewidth]{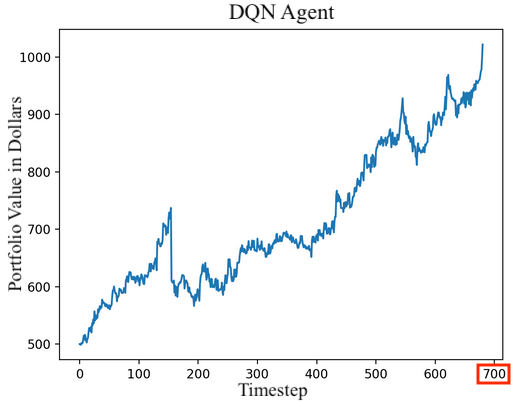}}
    \subfigure{\includegraphics[width=.246\linewidth]{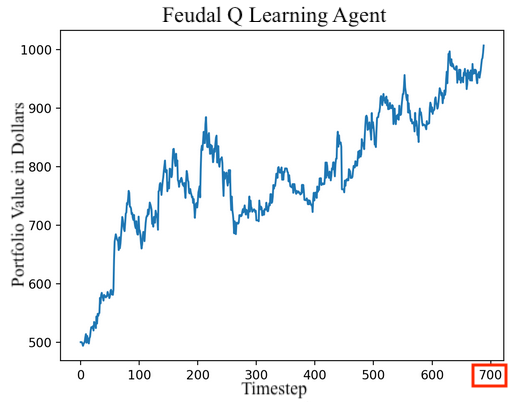}}
    \caption{Comparison of hard coded, q learning, DQN, and feudal q learning agent performance, respectively, in the task of doubling the value of a portfolio in the stock market. The hard coded agent is the slowest, followed by the q learning agent. The DQN agent and the feudal reinforcement learning agent performed comparably. Note the difference in the x-axis scales.}
    \label{fig:stockAgentComp}
\end{figure*}

\begin{figure*}[]
    \centering
    \subfigure{\includegraphics[width=.25\linewidth]{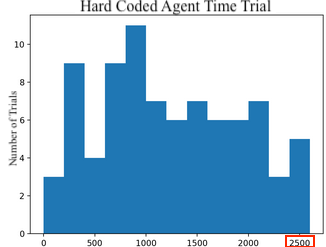}}
    \subfigure{\includegraphics[width=.247\linewidth]{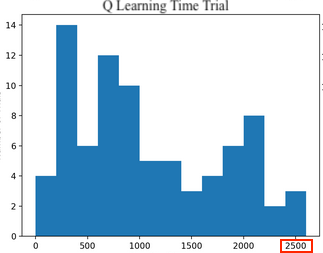}}
    \subfigure{\includegraphics[width=.243\linewidth]{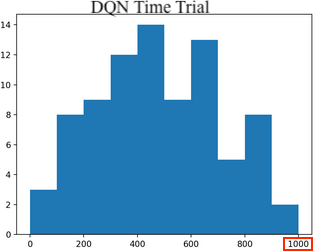}}
    \subfigure{\includegraphics[width=.246\linewidth]{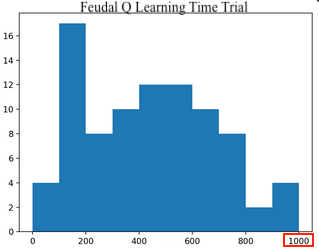}}
    \caption{We run the experiment from Figure \ref{fig:stockAgentComp} multiple times and record the optimal solution durations from each trial in histograms. Notice the difference in time scales between the first two and the last two histograms. The ranking from the previous figure still holds. However, feudal reinforcement learning has cemented itself as the fastest method, as evidenced by the fact that it's histogram is skewed to the left more than the DQN's.}
    \label{fig:portfolioTimeTrials}
\end{figure*}

Figure \ref{fig:feudalHigher} shows the reward per episode for each of the three maze agents. The reward of the three agents converges to the same number, by design, but it is clear from the graph that the feudal agent with a quadrant as the goal vector performs the best. Its reward reaches the convergence value much faster than either of the other two agents. However, both of the feudal agents have a large dip in the reward early on in training that is not present in the q learning agent, indicating that the feudal networks do a lot more of their exploration in the earlier stages of their training than the q learning network. 

\subsection{Portfolio Stock Experiments}
Now that we've discovered the power of feudal reinforcement learning, we revisit the problem of predicting stock prices. However, instead of attempting to solve a regression problem, we shift our focus to learning a policy over actions. We compare the performance of a hard coded agent, a q learning agent, a DQN agent, and a feudal Q learning agent at the task of doubling the value of a given portfolio in the stock market in Figure \ref{fig:stockAgentComp}. Pay attention to the x-axes in this figure to note the difference between the presented methods. The hard coded agent takes the longest amount of time to accomplish this task, as expected, with a duration of 1453 time steps. Also unsurprisingly, the q learning agent has the next longest duration with 802 time steps. The DQN agent doubles its portfolio value in 679 time steps, and the feudal q learning agent achieves this goal in 680 time steps. 

The main takeaway from this result is that we achieved comparable results with feudal q learning, which is a relatively simple method, as we did with the DQN, which is a relatively complicated, deep learning method. We wanted to verify that this was always the case, so we repeated this portfolio doubling experiment multiple times for each agent and recorded the duration results in the histograms in Figure \ref{fig:portfolioTimeTrials}. The y-axis is the number of trials in each bin, and the x-axis is the duration of each trial. Extra attention should be paid to the x-axes of the graphs in Figure \ref{fig:portfolioTimeTrials}. The hard coded and q learning agents have an x-axis from 0 to 2500, while the other two agent's x-axes are capped at 1000.

We can see that the q learning agent has values skewed more towards zero than the hard coded agent, so we expect an overall faster average duration from that agent. The average duration for the hard coded agent was 1521 time steps, and the q learning agent had an average duration of 1326 time steps, so this we prove this claim to be correct. In the same way, the feudal q learning agent has a histogram that is skewed more towards zero than the DQN agent, so we expect this to be the faster method. The DQN agent took an average of 651 time steps, and the feudal q learning agent took an average of 573 time steps. Therefore, we show that our original result was an understatement, and feudal q learning is, on average, much faster than a DQN at doubling a portfolio's value in the stock market. 

\subsection{Steering Angle Experiments}
\begin{figure}[]
    \centering
    \includegraphics[width=.9\linewidth]{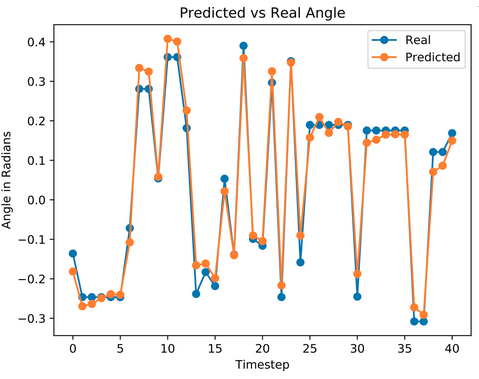}
    \caption{Steering angle predictions on the Udacity dataset from the modified Udacity steering challenge winner network. Note that this network takes the previous angle as input when predicting the next angle which gives it an unfair advantage.}
    \label{fig:komanda}
\end{figure}

In the stock portfolio experiments, we prove the effectiveness of feudal reinforcement learning. In this section, we aim to explore the boundary of its abilities in the driving domain. Our first experiment involves predicting steering angles based on image input. We create an image cube with ten sequential frames that we feed into our modified Udacity challenge network\cite{komanda}, along with the previous steering angle, to predict the next steering angle. Figure \ref{fig:komanda} shows a subset of real steering angles from the Udacity\cite{udacity} dataset, in blue, and the corresponding predicted angles, in orange. The predictions follow the real angles very closely except when there are drastic changes in the steering angles where it tends to over/under estimate the steering angle, which is the same issue we encountered with the LSTM stock experiments. 

\begin{figure*}[]
    \centering
    \subfigure{\includegraphics[width=.47\linewidth]{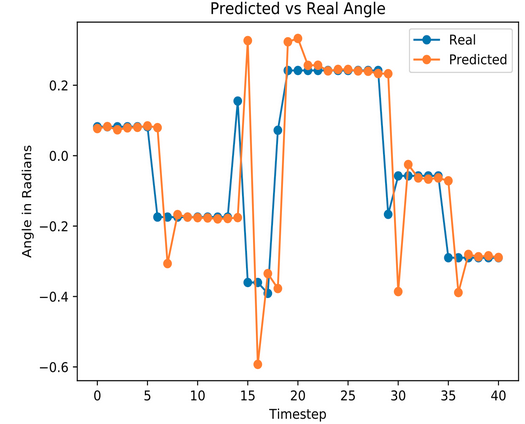}}
    \subfigure{\includegraphics[width=.47\linewidth]{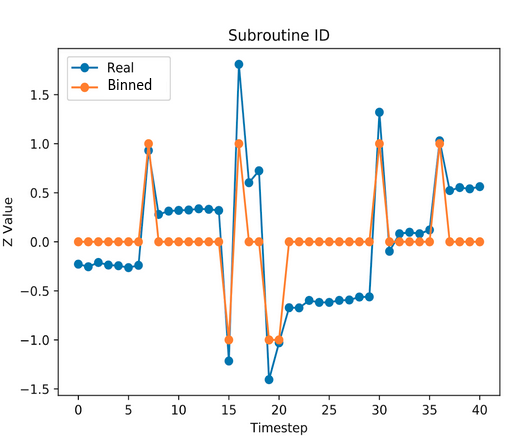}}
    \caption{Angle and subroutine ID prediction results on the Udacity dataset. Notice that the subroutine ID's behavior mimics the real angle behavior.}
    \label{fig:predZK}
\end{figure*}

Our ultimate goal, however, is to use feudal networks to predict steering angles. To do this, we first need to label subroutines within the data in order to have data with which to train the manager network. Instead of doing this by hand, we jointly train two networks: one that takes in a sequence of angles and predicts their subroutine ID, and another that takes in this subroutine ID, an image cube, and the previously predicted angle and predicts the next angle in the sequence. Figure \ref{fig:predZK} shows these prediction results. The left graph contains the steering angle predictions. The real angles are in blue, and the predicted angles are in orange. The right graph shows the predicted subroutine IDs. The blue line is the raw prediction values, and the orange line shows the binned values. For this, we map the predicted subroutine IDs to their closest value in the set $\{-1, 0, 1\}$. In this way, we have three discrete subroutine IDs corresponding to left turns, right turns, and going straight.

However, there are two problems with these solutions. The first is that it stands to reason that there could be more than just three subroutines represented in the driving data. Driving is a complex task that involves a lot of minutia. For instance, we could expand left turns to turning a little left, turning a moderate amount of left, and turning a lot left. The same could be done for right turns and even going straight. Therefore, constraining the subroutine IDs to fit into three discrete categories, as inspired by \cite{kumar2019learning}, may not allow us to represent an agent's actions thoroughly enough. The second problem is that, ideally, we want a network that predicts steering angles without explicitly taking in information about the previous angle because this gives the network an unfair advantage.

\begin{figure}[]
    \centering
    \subfigure{\includegraphics[width=.9\linewidth]{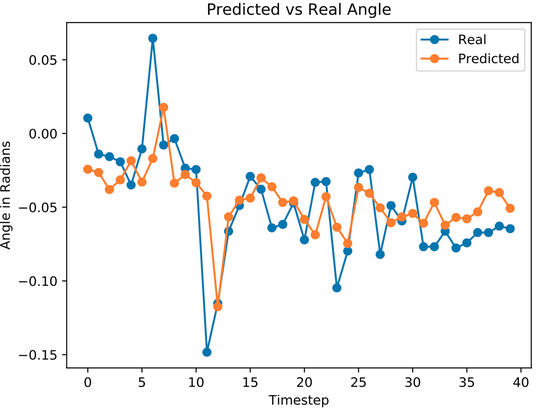}}
    
    \subfigure{\includegraphics[width=.9\linewidth]{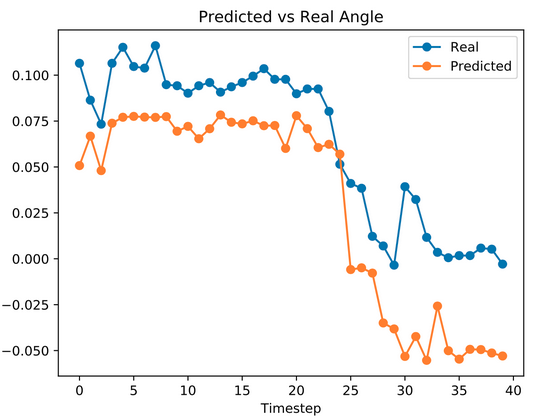}}
    \caption{Results of steering angle prediction when the t-SNE coordinates of the input data are used as the subroutine IDs. Notice that, for these results, we use a network that does not take the previous angle as input.}
    \label{fig:tsneSubIDAngles}
\end{figure}

To this end, we shift our focus from handcrafting our subroutine ID definitions to using t-SNE to do it automatically. We embed the data into 2D space and use those coordinate pairs as the subroutine IDs. However, before we attempt to predict the t-SNE coordinates from image data, we run an experiment to determine if the t-SNE coordinates will work as subroutine IDs. We use the ground truth value of the t-SNE centroids as the subroutine ID in our angle prediction network, along with an image cube of size ten, to determine whether or not it would be worthwhile to attempt to predict the centroids. If using t-SNE as the subroutine ID produces inaccurate results, then we would need to explore other avenues. The results of this are in Figure \ref{fig:tsneSubIDAngles}. The blue lines are the real steering angle, and the orange lines are the predicted angle. While the results in this figure are less accurate than our other prediction results, the predictions are more relevant to real world applications because they are computed using only visual input. 

\section{Discussion}
% Main Points:
% \begin{itemize}
    % \item LSTM \textrightarrow Multiplier/hard coded \textrightarrow q learning \textrightarrow DQN \textrightarrow feudal q learning \textrightarrow deep feudal
    % \item for: stock \textrightarrow maze \textrightarrow stock \textrightarrow driving
    % \item
%     \item LSTMs can't deal with quick, local changes but RL helps to alleviate that (although not completely)
    % \item feudal is faster in training than RL and does more exploration early on in training than RL (by maze)
    % \item feudal q achieves same results/better than DQN (by stocks)
    % \item tsne as subroutine ID produces adequate results even given the shift in the centroid and does so without previous angle!!!!
% \end{itemize}

In this work, we show that feudal reinforcement learning is more effective than reinforcement learning at the tasks of stock price prediction and steering angle prediction. We originally considered using generative adversarial networks (GANs) for sequence prediction, but our early experiments pointed to the effectiveness of feudal reinforcement learning instead. With our maze experiments, we find that feudal reinforcement learning is faster during training than reinforcement learning. We also find that feudal reinforcement learning achieves the maximum reward more quickly than reinforcement learning. Both of these effects are due to feudal reinforcement learning's temporal abstraction. Breaking down the problem into more easily digestible pieces narrows the focus of the worker agent and allows the optimal policy to be found more quickly.  

Additionally, temporal abstraction also helps alleviate the problems of long term credit assignment and sparse reward signals. The lower temporal resolution of the manager shortens the period of time between rewards overall. In addition to the original sparse reward, the worker network also receives a reward for obeying the goals from the manager. This feedback can be much more frequent than the sparse reward, thus allowing for more consistent network updates. In this way, we were able to achieve better results with feudal q learning in our stock portfolio experiments than with a DQN. 

Finally, we find that a t-SNE embedding space can be useful as the goal space for the manager in feudal reinforcement learning in our steering angle prediction experiment. We use the centroid corresponding to steering angle, braking, and throttle data from the previous ten time steps as the subroutine ID in our angle prediction network and were able to predict future steering angles without the direct use of the steering angle from the previous time step. The temporal abstraction inherent in the t-SNE centroid creation mimics the role of the manager network and allows the worker to be able to more accurately predict steering angles than if it attempted this task on its own.

{\small
\bibliographystyle{ieee_fullname}
\bibliography{ref}

\begin{thebibliography}{10}\itemsep=-1pt

\bibitem{mazeEnv}
Matthew Chan.
\newblock gym-maze.
\newblock \url{https://github.com/MattChanTK/gym-maze}, 2017.

\bibitem{chollet2015keras}
Fran\c{c}ois Chollet et~al.
\newblock Keras.
\newblock \url{https://keras.io}, 2015.

\bibitem{dayan1993feudal}
Peter Dayan and Geoffrey~E Hinton.
\newblock Feudal reinforcement learning.
\newblock In {\em Advances in neural information processing systems}, pages
  271--278, 1993.

\bibitem{komanda}
Komanda.
\newblock
  \url{https://github.com/udacity/self-driving-car/blob/master/steering-models/community-models/komanda/solution-komanda.ipynb},
  2016.

\bibitem{kumar2019learning}
Ashish Kumar, Saurabh Gupta, and Jitendra Malik.
\newblock Learning navigation subroutines by watching videos.
\newblock {\em arXiv preprint arXiv:1905.12612}, 2019.

\bibitem{maaten2008visualizing}
Laurens van~der Maaten and Geoffrey Hinton.
\newblock Visualizing data using t-sne.
\newblock {\em Journal of machine learning research}, 9(Nov):2579--2605, 2008.

\bibitem{kaggleStocks}
Boris Marjanovic.
\newblock Huge stock market dataset.
\newblock
  \url{https://www.kaggle.com/borismarjanovic/price-volume-data-for-all-us-stocks-etfs},
  Nov 2017.

\bibitem{mnih2013playing}
Volodymyr Mnih, Koray Kavukcuoglu, David Silver, Alex Graves, Ioannis
  Antonoglou, Daan Wierstra, and Martin Riedmiller.
\newblock Playing atari with deep reinforcement learning.
\newblock {\em arXiv preprint arXiv:1312.5602}, 2013.

\bibitem{QuandlWIKI}
Quandl.
\newblock {WIKI} various end-of-day data.
\newblock \url{https://www.quandl.com/data/WIKI}, 2016.

\bibitem{silver2017mastering}
David Silver, Thomas Hubert, Julian Schrittwieser, Ioannis Antonoglou, Matthew
  Lai, Arthur Guez, Marc Lanctot, Laurent Sifre, Dharshan Kumaran, Thore
  Graepel, et~al.
\newblock Mastering chess and shogi by self-play with a general reinforcement
  learning algorithm.
\newblock {\em arXiv preprint arXiv:1712.01815}, 2017.

\bibitem{udacity}
Udacity.
\newblock Udacity self-driving car driving data 10/3/2016
  (dataset-2-2.bag.tar.gz).

\bibitem{vezhnevets2017feudal}
Alexander~Sasha Vezhnevets, Simon Osindero, Tom Schaul, Nicolas Heess, Max
  Jaderberg, David Silver, and Koray Kavukcuoglu.
\newblock Feudal networks for hierarchical reinforcement learning.
\newblock In {\em Proceedings of the 34th International Conference on Machine
  Learning-Volume 70}, pages 3540--3549. JMLR. org, 2017.

\end{thebibliography}
}

\end{document}